\documentclass{Interspeech2024}

\usepackage{tabularray}
\usepackage{multirow}

\interspeechcameraready

\title{Transferable speech-to-text large language model alignment module}

\name[affiliation={1}]{Boyong}{Wu}
\name[affiliation={1}]{Chao}{Yan}
\name[affiliation={1}]{Haoran}{Pu}

\address{
  $^1$Cloudwalk Technology, China
}
\email{wuboyong@cloudwalk.com, yanchao@cloudwalk.com, puhaoran@cloudwalk.com}


\begin{document}

\maketitle

\begin{abstract}
    
    By leveraging the power of Large Language Models(LLMs) and speech foundation models, state of the art speech-text bimodal works can achieve challenging tasks like spoken translation(ST) and question answering(SQA) altogether with much simpler architectures. In this paper, we utilize the capability of Whisper encoder and pre-trained Yi-6B. Empirical results reveal that modal alignment can be achieved with one layer module and hundred hours of speech-text multitask corpus. We further swap the Yi-6B with human preferences aligned version of Yi-6B-Chat during inference, and discover that the alignment capability is applicable as well. In addition, the alignment subspace revealed by singular value decomposition(SVD) also implies linear alignment subspace is sparse, which leaves the possibility to concatenate other features like voice-print or video to expand modality.
\end{abstract}

\section{Introduction}
LLMs have received much attention in recent years. The powerful capabilities of ChatGPT\cite{10.5555/3495724.3495883} have achieved unprecedented breakthroughs in the natural language processing(NLP) field. Gradually, using a single model to solve multiple tasks has become the mainstream approach. Vision large language models have applied this principle to various vision tasks\cite{NEURIPS2022_960a172b, pmlr-v202-li23q, NEURIPS2023_6dcf277e, pmlr-v139-radford21a, zhu2023minigpt4}. In terms of speech modality, some studies have signaled that it is feasible to interact with LLM through speech. AudioGPT\cite{huang2023audiogpt} and HuggingGPT\cite{NEURIPS2023_77c33e6a} have made preliminary attempts. They employ a cascade method to seamlessly integrate automatic speech recognition (ASR), text-to-speech (TTS), and other recognition/generation tasks. The key concept is to apply LLM as an intermediate interface for distributing tasks via calling upon the appropriate models. 
Because the LLM is trained with text, speech information is hardly recognized, such as emotions and tones in human voice. By discretizing the speech signal into token sequences and expanding them within the LLM, SpeechGPT\cite{zhang-etal-2023-speechgpt} enables seamless text-speech interaction with a vocoder model for speech synthesis. However, this method requires retraining the LLM to support additional tokens. Moreover, there are some works achieve similar results by concatenating speech and text features as the prompt of LLM. LLaSM\cite{shu2023llasm} uses Whisper\cite{pmlr-v202-radford23a} and Chinese-LLAMA2-7B\footnote{https://huggingface.co/LinkSoul/Chinese-Llama-2-7b} as speech encoder and LLM with two training stages. In the first stage, they use ASR dataset for the adaptor pre-training. In the second stage, adaptor and language model are updated for cross-modal instruction fine-tuning. Whisper does not appear in Speech-LLaMA\cite{10389705}, they train 4 Transformer layers as audio encoder to complete ST tasks in 13 languages with LLaMA\cite{touvron2023llama}. Whisper and Qwen\footnote{https://huggingface.co/Qwen/Qwen1.5-7B} are used in Qwen-Audio\cite{chu2023qwenaudio}, which is also trained in two stages. The first stage, Qwen LLm is frozen and multi-task audio data is used to train Whisper. In the second stage of training, multi-round dialogue data is utilized to generate an interactive chat model that can accommodate input from diverse audio and text sources.

Previous works have excelled in aligning speech text modalities, most of which require retraining the speech encoder with a large amount of data to improve representation ability, and then fine-tuning the LLM model with instruction data to achieve better performance. However, this brings a large overhead to computing resources and is difficult to implement when data resources are scarce. Besides, those training strategies are fixed with particulay models and require multiple training with different speech-text foundation models composition. Should a replacement become necessary, realignment processes would have to be updated once more, leading to significant expenses in terms of overall training and utilization.

This paper raises several questions in response to this situation. Does modal alignment module require retraining speech modality and text modality? Does the alignment of speech-text modality only require a simple alignment module, or even a simple linear layer? Does training modality alignment module require massive amounts of data? Is the trained alignment module scalable and can it be replaced by a LLM with better performance? Furthermore, what kind of knowledge does the feature after alignment module mapping contain? Extend from existing work, we investigate the sufficiency of each components separately, namely the model size of the alignment module, amount of training data, transferability of alignment module across LLMs and the information contained in alignment modules which are rarely explored in current work.

\begin{figure*}[tb]
  \centering
  \includegraphics[width=0.8\textwidth]{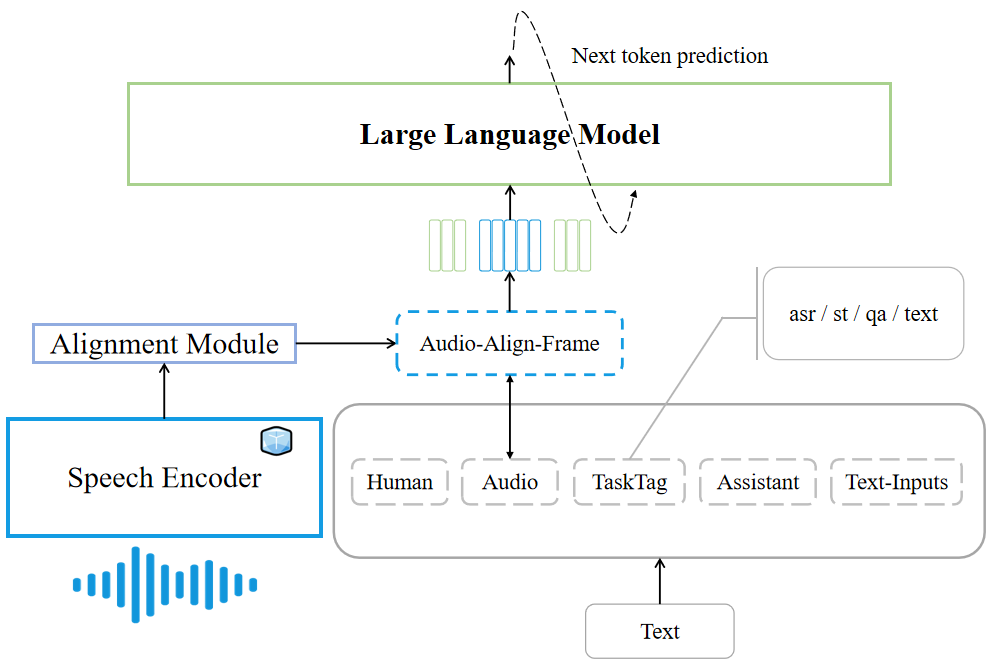}
  \caption{An overview of our proposed speech-text bimodal architecture. Alignment module is used to map the speech features into text feature space. Speech encoder is frozen all the time. LLM embedding will extract text features form prompt. The speech and text modal features are concatenated as LLM's input.}
  \label{fig:model}
\end{figure*}
We propose a linear layer after speech encoder as modal alignment module with open source models and corpus to achieve ASR, ST, SQA and text question answering(QA) multitasks in Mandarin. First, we conjecture that pre-trained speech encoder and LLM have strong text and speech capabilities, so we explore the connection through a single-layer alignment module. We choose Whisper encoder to extract speech features, while keeping parameters frozen, in order to reduce training overhead. Yi-6B\footnote{https://huggingface.co/01-ai/Yi-6B} with the LLaMA\cite{touvron2023llama} decoder-only structure is selected as LLM. A linear layer is chosen as the modal alignment module to map the speech features output by Whisper into the text feature space. The LLM is frozen during the alignment module training phase. In addition, we explore the transferability of alignment module. After the alignment module is aligned between speech and text, both of Whisper and alignment module are frozen, we replace the Yi-6B model with a supervised fine-tuning(SFT) version that aligns with human preferences. This updated model is validated on ST tasks, resulting in significant performance improvements. Finally, in order to further explore the alignment subspace, we use SVD analysis and therefore reveal information redundancy.  Our contributions can be summarized as the following points: 

\begin{itemize}
\item  Only adding and training an additional layer of alignment module between LLM and speech encoder to achieve ASR, ST, SQA and QA via open source models and data. The alignment module uses only a small amount of data to stimulate modal alignment capabilities.

\item The trained alignment module has strong scalability. It can be replaced with the SFT model with better command following and human preference capabilities from the same source without additional training, further improving the preference of specific tasks, such as ST, SQA, etc.

\item Preliminary analysis of the features after alignment mapping revealed information redundancy. Gradually reducing the dimension of modal alignment mapping revealed that a small reduction in feature dimension has only a slight impact on model performance. This provides insights for future feature concatenation, such as voiceprint features or video features.
\end{itemize}

\section{Approach and Experiment Setup}

\subsection{Model Architecture}
\textbf{Figure ~\ref{fig:model}} shows the model structure, including speech encoder, modal alignment module and LLM. Given the paired data $(s,x)$, where $s$ and $x$ denote the features mapped by alignment module and the text features extracted by LLM's embedding layer respectively. The training objective is to maximize the next token probability as $$P_{\theta}(x_t|x_{\textless t}, Alignment_{\phi}(s)),$$ where $\theta$ and $\phi$ denote the parameters of the large language model and the alignment module.

\noindent\textbf{Speech Encoder} The speech encoder uses the encoder module of Whisper large-v3\footnote{https://huggingface.co/openai/whisper-large-v3}, which is trained on 1 million hours of weakly labeled audio and 4 million hours of pseudolabeded audio collected using Whisper large-v2\cite{pmlr-v202-radford23a}. The encoder module accepts a 128-dimensional mel-spectrogram as input and produces an output with a dimension of 1280.

\noindent\textbf{Large Language Model} The open source Yi-6B is selected as the LLM. It is a bilingual language model that supports Chinese and English, which is trained on a 3T multi-language corpus. Yi-6B is a 32-layer transformer decoder-only structure with a hidden size of 4096. 

\noindent\textbf{Alignment Module} Modal alignment module is a linear layer that maps the features output by the Whisper encoder to the LLM text modality, with an input dimension of 1280 and an output dimension of 4096.

\begin{figure}[t]
  \centering
  \includegraphics[width=\linewidth]{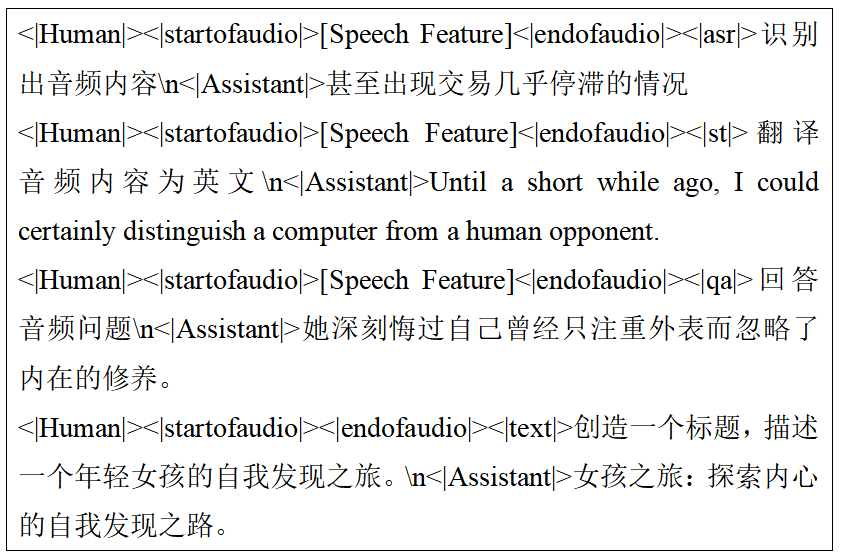}
  \caption{Cases of speech and plain text input}
  \label{fig:use_cases}
\end{figure}

\subsection{Prompt design}
Since a speech-text bimodal LLM requires support for both audio and text inputs, we design the data format inspired from Whisper\cite{pmlr-v202-radford23a} and Qwen-Audio\cite{chu2023qwenaudio}. For the input sequence, ``\textless\textbar Human\textbar\textgreater'' is the special token, which means that the following content from here on is provided by humans. The next special token is ``\textless\textbar startofaudio\textbar\textgreater'',  the audio content will be connected after this token. And then, special token ``\textless\textbar endofaudio\textbar\textgreater'' is followed, which represents the end of the speech content. In cases where the input lacks speech content, the area enclosed by ``\textless\textbar startofaudio\textbar\textgreater'' and ``\textless\textbar endofaudio\textbar\textgreater'' will be empty. The next special token is \{task\}, which is used to specify the model generation task and the \{prompt\} will follow it for LLM. Different tasks have unique \{prompt\}. For the ASR task, the prompt is to ``recognize the content in the speech". ``Translate audio content into English" is for ST. The prompt for QA is ``Answer the question in the audio". And the plain text task takes its prompt from the question presented in the description. After \{prompt\}, the special token "\textless\textbar Assistant\textbar\textgreater" indicates that the subsequent content generated by the model is the label content of the current sample. \textbf{Figure ~\ref{fig:use_cases}} shows the cases with speech and plain text.

\subsection{Training strategy}
The training process requires the following steps. We first extract the fbank feature from the audio data via Whisper's default configuration and generate speech feature by speech encoder. Then the speech feature passes through the alignment module and concatenate with LLM text embedding of the prompt and start of the answer.  Finally, the alignment module will be optimized by CrossEntropy loss.

\subsubsection{Modal alignment}
\label{section:modal_alignment}
Whisper has excellent performance in ASR and ST tasks, and its encoder has strong semantic representation capabilities. As a LLM base model, Yi-6B demonstrates robust language capabilities due to its extensive pre-training using vast amounts of text data. Given the rich representations from text and speech foundation models, we explore the possibility to achieve modality alignment via a single linear layer. In this stage of training, freeze the parameters of the speech encoder and LLM, and use the ASR, ST, and SQA data to train the modal alignment module. We will explore how much data the modal alignment module requires to stimulate modal capabilities.


\subsubsection{Extensibility of the alignment module}
We also explore whether the alignment module has transfer-ability. Based on \ref{section:modal_alignment}, we keep speech encoder and alignment module untouched, while swap the original LLM model with a homologous SFT model with stronger instruction following and human preference capabilities, and test it against ST data.

\subsubsection{Alignment mapping feature analysis}
In order to analyze the feature content of the alignment module after mapping, we use the SVD algorithm to perform feature decomposition.
$$A_{m*n}=U_{m*m}\Sigma_{m*n}V_{n*n}^{T}$$
$A$ is the speech feature matrix mapped by the modal alignment module. $m$ and $n$ are the time dimension and the hidden size of the LLM respectively. $\Sigma$ is an $m*n$ matrix, all of which are 0 except for the elements on the main diagonal. The main diagonal each element on the line becomes a singular value, and elements closer to the top are more important. We will only retain the top part the $\Sigma$ value to explore whether erasing feature information will have a greater impact on model performance.

\subsection{Experiments setup}
\subsubsection{Experimental data}
For the speech recognition task, we use the open source aishell\cite{8384449} and WenetSpeech\cite{9746682} data. For the translation task, we use the wmt19\cite{Ng2019FacebookFW} Chinese-English data set, with a total of 310k items. For the QA task, we use the Alpaca-zh\cite{wang-etal-2023-self-instruct} dataset, which has a total of 48k pieces of data. Both ST and SQA tasks use a self-developed TTS model to generate audio from text data, and randomly select speaker information to ensure the diversity of speech timbres. There are also many excellent open source TTS models available, such as Bark-TTS\footnote{https://github.com/suno-ai/bark}, etc. The synthesized data is about 643 hours for wmt19 and about 60 hours for Alpaca.

\noindent We build the following multi-task dataset:
\begin{itemize}
\item dataset1. 90 hours aishell, 100 hours wmt19 and 30 hours Alpaca.
\item dataset2. 178 hours aishell, 200 hours wmt19 and 60 hours Alpaca.
\item dataset3. 178 hours aishell, 200 hours WenetSpeech, 200 hours wmt19 and 60 hours Alpaca.
\end{itemize}
For the test set, ASR task uses the test set of aishell2\cite{du2018aishell2}. ST task uses the test set of wmt19.
\subsubsection{Parameter settings}
The Speech encoder uses the Whisper's encoder module of large-v3, the LLM uses the Yi-6B, and the modal alignment module uses only a linear layer.

When training the modal alignment module, freeze both the speech encoder and LLM parameters, only update the parameters of modal alignment module . Learning rate is set to 1e-3, batch size is set to 128, and A800-40G is used for training.

LoRA\cite{hu2022lora} is used when fine-tuning LLM, the speech encoder and alignment module are frozen. Learning rate is set to 1e-4, and the batch size is set to 128. For LoRA parameters, $r$ is set to 16 and $\alpha$ is set to 32.

For audio, all the data are \SI{16}{\kilo\hertz} single-channel in wav format. The fBank feature uses \SI{25}{\milli\second} window size and a hop size of \SI{10}{\milli\second}.

As for evaluation, the ASR task uses the character error rate(CER) as the statistical standard, and ROUGE-L for ST task.

\section{Results and Analysis}

\begin{table}[]
\centering
\caption{ROUGE-L (\%) and CER (\%) score for alignment module experiments for Yi-6B and Yi-6B-Chat. CER are evaluated by AISHELL-2 and ROUGE-L score are evaluated by WM19}
\label{tab:evaluation}
\begin{tabular}{|c|c|c|c|c|}
\hline
\multirow{2}{*}{} & \multirow{2}{*}{LLM} & \multicolumn{3}{c|}{Training Dataset} \\ \cline{3-5} 
  &  & 1 & 2 & 3 \\ \hline
  
\multirow{3}{*}{ROUGE-L ($\uparrow$)}
 &  Yi-6B & 29.378 & 31.392 & 27.916  \\ \cline{2-5} 
 &  Yi-6B-Chat & 33.180 & \textbf{33.844} & 30.660 \\ \cline{2-5} 
 &  Yi-6B-LoRA & 29.697 & 31.512 & 27.877 \\ \hline
 
\multirow{3}{*}{CER ($\downarrow$)}
 &  Yi-6B & 11.071 & 9.418 & 8.429 \\ \cline{2-5} 
 &  Yi-6B-Chat & 12.753 & 14.824 & 8.616 \\ \cline{2-5} 
 &  Yi-6B-LoRA & 11.021 & 9.321 & \textbf{8.251} \\ \hline

\end{tabular}
\end{table}

In this section, we first go through evaluation result in speech recognition and translation. And then we perform a deeper analysis about how alignment module behaves during inference. 
\subsection{Evaluation}

\textbf{Table ~\ref{tab:evaluation}} of the Yi-6B section shows our alignment methods against three dataset configurations. All results are compared internally in order to investigate how linear alignment module behaves across different corpus size and composition, suggesting our method achieves text-speech modality alignment. For CER, we observe an incremental improvements of metrics from $11.071$ to $8.429$. On the other hand, alignment module trained from dataset 2 has the best ROUGE-L score ($31.392$) while there is a $3.476$ decreases from dataset 3 setting ($27.916$), worse than the result from dataset 1 ($29.738$), even though it contains much more speech data. The first observation suggests that adding training data for one task can indeed enhance the corresponding learnability, while the unbalanced data  significantly degenerates the tasks with minor utterances. Hundreds of hours of audio data can inspire the alignment capabilities of the module with only one linear layer. Beyond the training result evaluation, we urge future modality-align works to take better attention towards balance construction of Speech-LLM alignment training.

\subsection{Alignment module's transfer-ability across LLMs}

After the alignment module is trained, we swap the Yi-6B model with Yi-6B-Chat\footnote{https://huggingface.co/01-ai/Yi-6B-Chat} model fine-tuned by human preferences dataset to investigate its behaviors across choices of LLMs. The comparison of \textbf{Table ~\ref{tab:evaluation}} across LLM suggests the alignment module can still align speech and text modalities, given the input LLM is fine-tuned with specific tasks. However, given LLM fine-tuned with chat prefers to generate  semantically related content from prompt, there is a significant improvement for speech translation task around 3.0 improvement of ROUGE-L score, while there is a non-negligible degeneration of speech recognition capability, which renders around 2.4 increase of character  error rate for alignment module trained by average. With closer examination we find most of the wrong transcription are semantically the same as reference text with wrong pronunciation. In addition, we use LoRA to fine-tune Yi-6B with the same data used by alignment module and we can observe a minimal improvement in model performance. We conjecture that, with more and balance dataset, invariance between alignment module and LLM may push future speech-text alignment module to become independent from speech encoder and language models. Comparing with other LoRA-based alignment technique, such approach can achieve one-for-all alignment free from fine-tuning different LLM variances. When speech data resources are scarce, we can focus on more easily accessible text modal data and SFT LLM to improve the performance of speech-text modality LLM on specific tasks.

\subsection{Alignment Feature Analysis}

\begin{table}[]
\centering
\caption{ROUGE-L (\%) and CER (\%) score for top-k SVD decomposition inference for training dataset 2 configuration.}
\label{tab:svd}
\begin{tabular}{|c|c|c|c|c|c|c|}
\hline
 \multirow{2}{*}{} & \multicolumn{6}{c|}{Top-k singular vecrors} \\ \cline{2-7} 
  & None & 1000 &  300 & 200 & 100 & 50 \\ \hline
ROUGE-L ($\uparrow$) & 31.4 & 31.3 & 31.4 & 30.0 & 3.89 & 0.31  \\ \hline
CER ($\downarrow$) & 9.42 & 9.61 &  9.45 & 9.60 & 11.7 & 70.8 \\ \hline
\end{tabular}
\end{table}

\begin{table}[]
\centering
\caption{ROUGE-L (\%) and CER (\%) score for alginment module with various trainable dimensions under training dataset 2 configuration.}
\label{tab:subspace}
\begin{tabular}{|c|c|c|c|c|}
\hline
 \multirow{2}{*}{} & \multicolumn{4}{c|}{Trainable dimension} \\ \cline{2-5} 
  & 4096 & 3072 &  2048 & 1024  \\ \hline
ROUGE-L ($\uparrow$) & 31.392 & 31.144 & 24.958 & 22.011   \\ \hline
CER ($\downarrow$) & 9.418 & 9.365 &  14.447 & 19.486  \\ \hline
\end{tabular}
\end{table}

We first take the top-k singular vector decomposition of the linear alignment module during inference to measure the information entailment in alignment space. \textbf{Table ~\ref{tab:svd}} reveals that there is negligible change of ROUGE-L and CER by applying the top-200 or more singular vectors while scores suddenly. On the other hand, shifting from top-200 to top 50 there is a significant drop from 9.60 to 70.8 for CER and from 30.0 to 0.31 for ROUGE-L. Such observation implies that the aligned space can have much fewer rank compared with full-size 4096 LLM subspace.  As a supplementary experiment, we constrain trainable alignment modules with $\{3072, 2048, 1024\}$ trainable dimension, and then fill the gap dimension with 0 during training. \textbf{Table ~\ref{tab:subspace}} shows that there is negligible change of CER ($9.418\rightarrow 9.365$) and ROUGE-L ($31.392\rightarrow 31.144$) scores when the training dimension  decrease from  4096 to 3072 while there is a significant drop from 3072 to 1024 ($9.365\rightarrow 19.486$ for CER and $31.144\rightarrow 22.011$ for ROUGE-L). Given each learnable dimension represents the highest possible ranks for alignment subspace, such observation further implies that alignment space might be less complicated compared with text subspace described by LLM which may lead further simplification.

\section{Conclusion}

We explore the capability of speech-text multitasking by training an linear alignment module across Whisper and Yi-6B models. Results from ASR and ST reveal that speech-text alignment module can be achieved and the balance of the dataset significantly impacts each tasks capability. In further, the extensibility across Yi-6B and Yi-6B-Chat version and the alignment module's sparse space also suggests its universal applicability and to extend with more tasks. Future work is required for investigating the further potential such as additionally integrating video as the third input or other acoustic related tasks.

\newpage

\bibliographystyle{IEEEtran}
\bibliography{mybib}

\end{document}